\documentclass[10pt,twocolumn,letterpaper]{article}

\usepackage{cvpr}              

\usepackage{graphicx}
\usepackage{amsmath}
\usepackage{amssymb}
\usepackage{booktabs}
\usepackage{algorithmic}
\usepackage{algorithm}

\usepackage{bm}
\usepackage{multirow}

\usepackage[pagebackref,breaklinks,colorlinks]{hyperref}

\usepackage[capitalize]{cleveref}
\crefname{section}{Sec.}{Secs.}
\Crefname{section}{Section}{Sections}
\Crefname{table}{Table}{Tables}
\crefname{table}{Tab.}{Tabs.}

\def\cvprPaperID{9169} 
\def\confName{CVPR}
\def\confYear{2023}

\begin{document}

\title{PCR: Proxy-based Contrastive Replay \\for Online Class-Incremental Continual Learning}

\author{Huiwei Lin, Baoquan Zhang, Shanshan Feng\thanks{Corresponding author}, Xutao Li, Yunming Ye\\
Harbin Institute of Technology, Shenzhen\\
{\tt\small \{linhuiwei, zhangbaoquan\}@stu.hit.edu.cn, \{victor\_fengss, lixutao, yeyunming\}@hit.edu.cn}
}
\maketitle

\begin{abstract}
Online class-incremental continual learning is a specific task of continual learning. It aims to continuously learn new classes from data stream and the samples of data stream are seen only once, which suffers from the catastrophic forgetting issue, i.e., forgetting historical knowledge of old classes.
Existing replay-based methods effectively alleviate this issue by saving and replaying part of old data in a proxy-based or contrastive-based replay manner. 
Although these two replay manners are effective, the former would incline to new classes due to class imbalance issues, and the latter is unstable and hard to converge because of the limited number of samples.
In this paper, we conduct a comprehensive analysis of these two replay manners and find that they can be complementary.
Inspired by this finding, we propose a novel replay-based method called proxy-based contrastive replay (PCR).
The key operation is to replace the contrastive samples of anchors with corresponding proxies in the  contrastive-based way. 
It alleviates the phenomenon of catastrophic forgetting by effectively addressing the imbalance issue, as well as keeps a faster convergence of the model.
We conduct extensive experiments on three real-world benchmark datasets, and empirical results consistently demonstrate the superiority of PCR over various state-of-the-art methods~\footnote{\url{https://github.com/FelixHuiweiLin/PCR}}.

\end{abstract}
\section{Introduction}
\label{sec:introduction}

Online class-incremental continual learning (online CICL) is a special scenario of continual learning~\cite{delange2021continual}. Its goal is to learn a deep model that can achieve knowledge accumulation of new classes and not forget information learned from old classes. In the meantime, the samples of a continuously non-stationary data stream are accessed only once during the learning process. At present, catastrophic forgetting (CF) is the main problem of online CICL. It is associated with the phenomenon that the model has a significant performance drop for old classes when learning new classes. The main reason is historical knowledge of old data would be overwritten by novel information of new data.

\begin{figure}[t]
\centering
\includegraphics[scale=0.48]{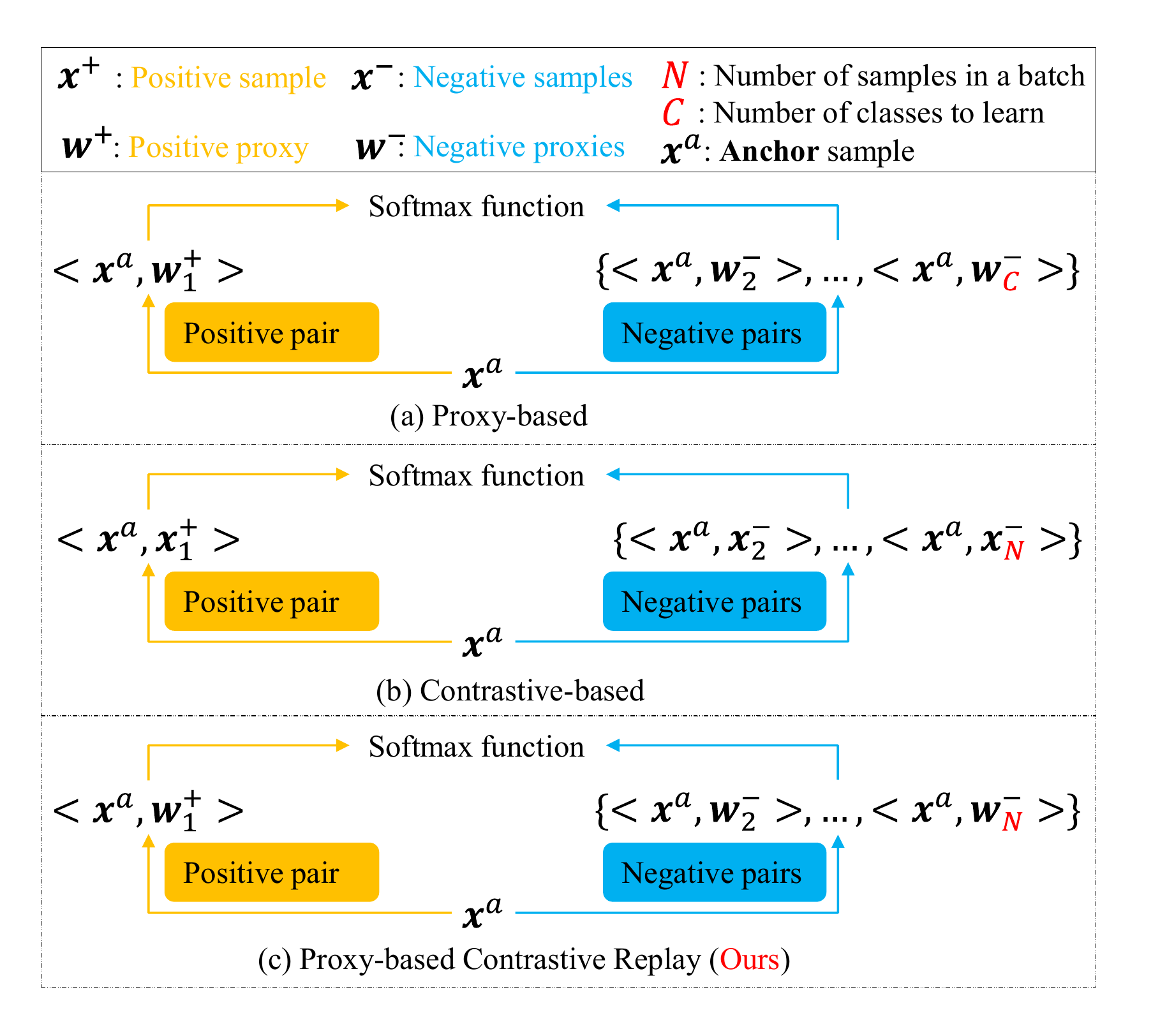}
\vspace{-1ex}
\caption{Illustration of our work. (a) The example of proxy-based replay manner. For each anchor sample, it calculates similarities of all anchor-to-proxy pairs. (b) The example of contrastive-based replay manner. For each anchor sample, it calculates similarities of all anchor-to-sample pairs in the same batch. (c) The example of our method. It calculates similarities of anchor-to-proxy pairs, which is similar to the proxy-based method. However, the anchor-to-proxy pairs are selected by the anchor-to-sample pairs in the same batch, which performs in the contrastive-based manner.}
\label{fig:illustration}%
\vspace{-3ex}
\end{figure}

Among all types of methods proposed in continual learning, the replay-based methods have shown superior performance for online CICL~\cite{mai2022online}. In this family of methods, part of previous samples are saved in an episodic memory buffer and then used to learn together with current samples. In general, there are two ways to replay. The first is the proxy-based replay manner, which is to replay by using the proxy-based loss and softmax classifier. As shown in Figure~\ref{fig:illustration}(a), it calculates similarities between each anchor with all proxies belonging to $C$ classes. A proxy can be regarded as the representative of a sub-dataset~\cite{yao2022pcl}, and the anchor is one of the samples in the training batch. The second is the contrastive-based replay manner that replays by using the contrastive-based loss and nearest class mean (NCM) classifier~\cite{mensink2013distance}. Shown as Figure~\ref{fig:illustration}(b), it computes similarities between each anchor with all $N$ samples in the same training batch. Although these two manners are effective, they have their corresponding limitations. The former is subjected to the ``bias'' issue caused by class imbalance, tending to classify most samples of old classes into new categories. The latter is unstable and hard to converge in the training process due to the small number of samples.

In this work, we comprehensively analyze their characteristics and find that the coupling of them can achieve complementary advantages. On the one hand, the proxy-based manner enables fast and reliable convergence with the help of proxies. On the other hand, although the contrastive-based manner is not very robust, it has advantages in the selection of anchor-to-sample pairs. Only the classes associated with samples in anchor-to-sample pairs can be selected to learn. Previous studies~\cite{ahn2021ss,caccia2021new} have proved that suitably selecting of anchor-to-proxy pairs is effective to address the ``bias'' issue. Therefore, it is necessary to develop a coupling manner to jointly keep these advantages at the same time. In other words, it not only takes proxies to improve the robustness of the model as proxy-based manner, but also overcomes the ``bias'' problem by selecting anchor-to-proxy pairs as the pairs selection of contrastive-based manner.

With these inspirations, we propose a novel replay-based method called proxy-based contrastive replay (PCR) to alleviate the phenomenon of CF for online CICL. The core motivation is the coupling of proxy-based and contrastive-based loss, and the key operation is to replace anchor-to-sample pairs with anchor-to-proxy pairs in the contrastive-based loss. As shown in Figure~\ref{fig:illustration}(c), our method calculates similarities between each anchor and other proxies, which is similar to the proxy-based loss. However, it does not straightly make full use of proxies from all classes. It only takes the proxies whose associated classes of samples appear in the same batch, which is analogous to the contrastive-based loss. For one thing, it keeps fast convergence and stable performance with the help of proxies. For another thing, it addresses the ``bias'' issue by only choosing part of anchor-to-proxy pairs to calculate categorical probability. And the selected anchor-to-proxy pairs are generally better than the ones selected by existing solutions~\cite{ahn2021ss,caccia2021new}.

Our main contributions can be summarized as follows:
\begin{itemize}
\item[1)] We theoretically analyze the characteristics of proxy-based and contrastive-based replay manner, discovering the coupling manner of them is beneficial. To the best of our knowledge, this work is the first one to combine these two manners for the online CICL problem.

\item[2)] We develop a novel online CICL framework called PCR to mitigate the forgetting problem. By replacing the samples for anchor with proxies in contrastive-based loss, we achieve the complementary advantages of two existing approaches.

\item[3)] We conduct extensive experiments on three real-world datasets, and the empirical results consistently demonstrate the superiority of our PCR over various state-of-the-art methods. We also investigate and analyze the benefits of each component by ablation studies.
\end{itemize}

\section{Related work}
\label{sec:relatedwork}
\subsection{Continual Learning}
Recent advances on continual learning are driven by three main directions. 
1) Architecture-based methods~\cite{zhu2021class}, also known as parameter-isolation methods, divide each task into a set of specific parameters of the model. They dynamically extend the model as the number of tasks increases~\cite{rusu2016progressive} or gradually freeze part of parameters to overcome the forgetting problem~\cite{miao2021continual}. 2) Regularization-based methods~\cite{zhu2021class}, also called prior-based methods, store previous knowledge learned from old data as prior information of network.  It takes the historical knowledge to consolidate past knowledge by extending the loss function with additional regularization term~\cite{kirkpatrick2017overcoming,dhar2019learning}. 3) Replay-based methods, which set a fixed-size memory buffer~\cite{lin2022anchor,lopez2017gradient,chaudhry2018efficient,farajtabar2020orthogonal,tang2021layerwise} or generative model~\cite{van2020brain,xiang2019incremental,cui2021deepcollaboration,choi2021dual} to store, produce, and replay historical samples in the training process, also go by the name rehearsal-based methods. This kind of methods~\cite{buzzega2020dark,cha2021co2l,chaudhry2021using,liu2021rmm,wang2021memory} that replay old samples in the buffer are still the most effective for anti-forgetting at present~\cite{buzzega2021rethinking}. 

\subsection{Online Class-Incremental Continual Learning.}
Replay-based methods based on experience replay (ER)~\cite{rolnick2019experience} are the main solutions of online CICL. Some approaches use the \textbf{memory retrieval strategy} to select valuable samples from memory, such as MIR~\cite{aljundi2019online} and ASER~\cite{shim2021online}. In the meantime, some approaches~\cite{aljundi2019gradient,jin2021gradient,he2021online} focus on saving more effective samples to the memory, belonging to the \textbf{memory update strategy}. The others~\cite{mai2021supervised,yin2021mitigating,caccia2021new,guo2022online,gu2022not} utilize the \textbf{model update strategy} to improve the learning efficiency. Recently, a new method AOP based on orthogonal projection has been proposed without buffer. Most of them are proxy-based manners except SCR~\cite{mai2021supervised}, which is a contrastive-based manner. 

The proposed PCR in this work exploits a new model update strategy for online CICL, belonging to the family of replay-based methods. Different from existing approaches, it aims to combine the contrastive-based replay manner with the proxy-based replay manner. By complementing their advantages, the coupling manner can more effectively alleviate the phenomenon of catastrophic forgetting.

\section{Problem Statement and Analysis}
\label{sec:preliminary}
\subsection{Problem Formulation} 
Online CICL divides a data stream into a sequence of learning tasks as $\mathcal{D}=\{\mathcal{D}_t\}_{t=1}^T$, where $\mathcal{D}_t=\{\mathcal{X}_t\times \mathcal{Y}_t,\mathcal{C}_t\}$ contains the samples $\mathcal{X}_t$, corresponding labels $\mathcal{Y}_t$, and task-specific classes $\mathcal{C}_t$. Different tasks have no overlap in the classes. The neural network is made up of a feature extractor $\bm{z}=h(\bm{x};\bm{\Phi})$ and a proxy-based classifier $f(\bm{z};\bm{W})=\langle\bm{z},\bm{W}\rangle\cdot\gamma$~\cite{hou2019learning}, where $\bm{W}$ contains trainable proxies of all classes, $\langle\cdot,\cdot\rangle$ is the cosine similarity, and $\gamma$ is a scale factor. All of learned classes are denoted as  $\mathcal{C}_{1:t}=\bigcup_{k=1}^t \mathcal{C}_k$. The categorical probability that sample $x$ belongs to class $c$ is
\begin{equation}
    \label{eq:probability}
	p_c=\frac{e^{\langle h(\bm{x};\bm{\Phi}),\bm{w}_c\rangle\cdot\gamma}}{\sum_{j\in\mathcal{C}_{1:t}}e^{\langle h(\bm{x};\bm{\Phi}),\bm{w}_j\rangle\cdot\gamma}}.
\end{equation}
In the training process, the model can only access ${D_t}$ and each sample can be seen only once. Its objective function is
\begin{equation}
    \label{eq:oclloss}
	L=E_{(\bm{x},y)\sim{\mathcal{D}_t}}[-log(\frac{e^{\langle h(\bm{x};\bm{\Phi}),\bm{w}_y\rangle\cdot\gamma}}{\sum_{j\in\mathcal{C}_{1:t}}e^{\langle h(\bm{x};\bm{\Phi}),\bm{w}_j\rangle\cdot\gamma}})].
\end{equation}

\subsection{Analysis of Catastrophic Forgetting.} 
A direct cause of CF is the unbalanced gradient propagation between old and new classes. The gradient for a single sample $\bm{x}$ can be expressed as

\begin{equation}
    \label{eq:gradient_h}
    \frac{\partial L}{\partial \bm{W}}=\left\{
    \begin{aligned}
        h(\bm{x};\bm{\Phi})(p_y-1)&,&i= y \\
        h(\bm{x};\bm{\Phi})(p_c)&,&c\neq y
    \end{aligned}
    \right..
\end{equation}

\noindent As Equation (\ref{eq:gradient_h}) shows, if a training sample $\bm{x}$ belongs to class $y$, it not only increases the logits value of the $y$-th dimension by $p_y-1<0$, but also decreases the logits value of other dimension by $p_c>0$. Combining with the chain rule, it provides the positive gradient for the proxy of class $y$ as $\bm{w}_y=\bm{w}_y-\eta h(\bm{x};\bm{\Phi})(p_y-1)$, and propagates the negative gradient to the other proxies as $\bm{w}_c=\bm{w}_c-\eta h(\bm{x};\bm{\Phi})(p_c)$. Since $\eta$ is a positive learning rate and $h(\bm{x};\bm{\Phi})$ is usually non-negative by Relu~\cite{nair2010rectified}. Furthermore, the gradient is transferred to the feature extractor, making it focus on the features that can distinguish this class from other classes.

\begin{figure*}[t]
\centering
\vspace{-1ex}
\includegraphics[scale=0.76]{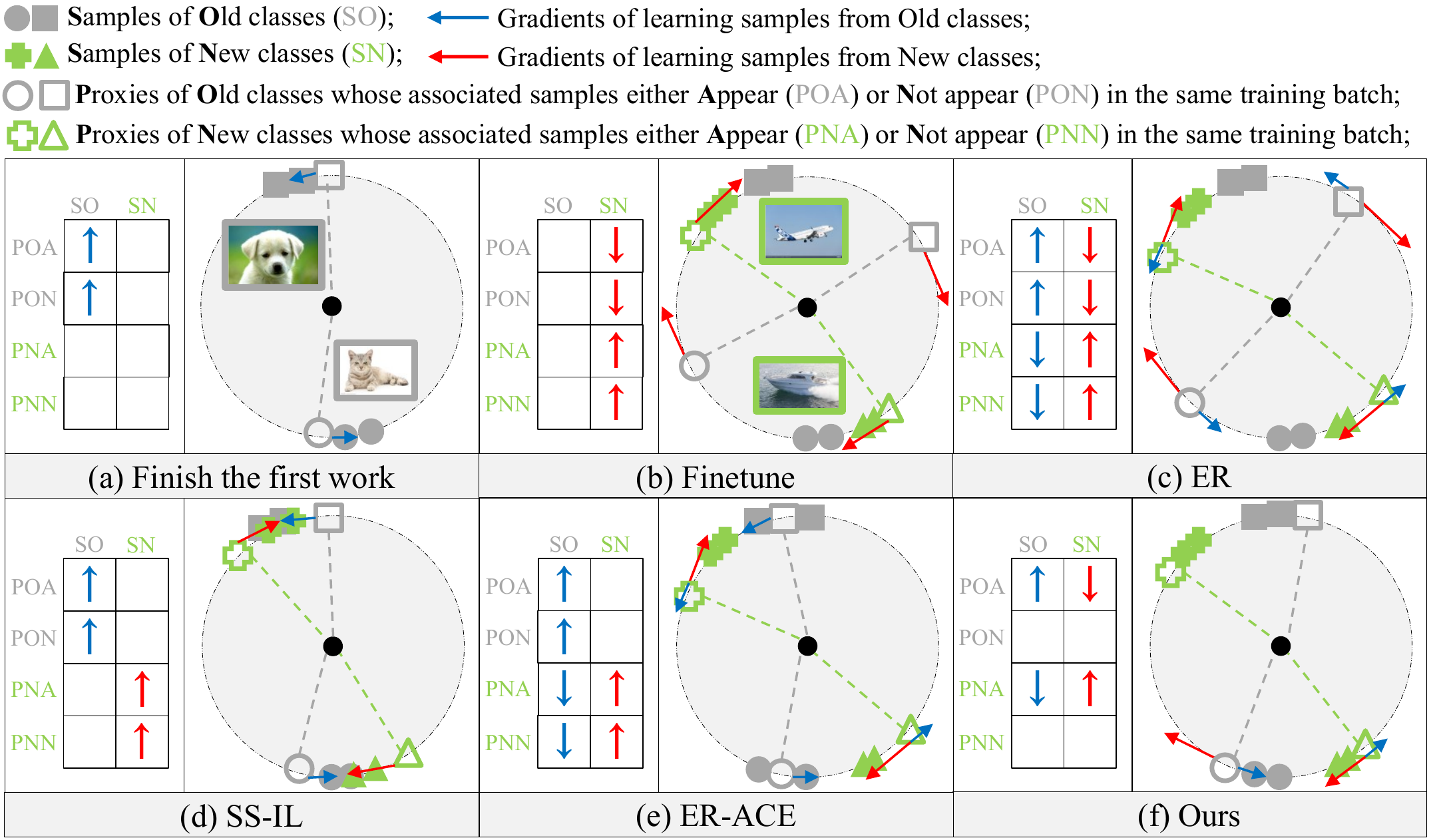}
\vspace{-2ex}
\caption{Analysis of existing proxy-based manners. In each sub-figure, the left part is the process of gradient propagation from samples to all proxies, and the right part is the unit embedding space of samples and proxies. (a) The learning of the first task. The gradient propagation only exists in current two classes, denoted as the blue arrows. (b) The learning of the second task by Finetune. The new classes dominate the gradient propagation, denoted as the red arrows. (c) The learning of the second task by ER. (d) The learning of the second task by SS-IL. (e) The learning of the second task by ER-ACE. (f) The learning of the second task by our method. Different from existing studies, our method controls the process of gradient propagation more effectively, improving the recognition of new and old classes.}
\label{fig:analysis}%
\vspace{-3ex}
\end{figure*}

When directly optimizing Equation (\ref{eq:oclloss}), which is known as Finetune, the learning of new classes dominates the gradient propagation, causing the phenomenon of CF. To better analyse it, we show a case that learns the samples of cat and dog at the first task (Figure~\ref{fig:analysis}(a)), and then learns the samples of ship and airplane at the next task (Figure~\ref{fig:analysis}(b)-(f)). As seen in the left part of Figure~\ref{fig:analysis}(b), the gradient is produced by learning new classes. As a result, the proxies of new classes receive more positive gradient ($\uparrow$) and the others obtain more negative gradient ($\downarrow$). Shown as the red arrows in the left part of Figure~\ref{fig:analysis}(b), it causes the proxies of new classes are close to the samples of new classes, while the proxies of old classes are far away from them. Meanwhile, the feature extractor pays more attention to the features of new classes. It causes the samples of new and old classes are close~\cite{caccia2021new} in the unit embedding space. Hence, it is easy to classify samples to new classes.

\subsection{Analysis of Proxy-based Manner}
\textbf{ER}~\cite{rolnick2019experience} allocates a memory buffer $\mathcal{M}$ to temporarily store part of previous samples of old classes, which are retrained with current samples. And its objective function is
\begin{equation}\small
\label{eq:erloss}
\begin{split}
    L_{ER}=E_{(\bm{x},y)\sim{\mathcal{D}_t\cup\mathcal{M}}}[-log(\frac{e^{\langle h(\bm{x};\bm{\Phi}),\bm{w}_y\rangle\cdot\gamma}}{\sum_{j\in\boxed{\mathcal{C}_{1:t}}}e^{\langle h(\bm{x};\bm{\Phi}),\bm{w}_j\rangle\cdot\gamma}})],
\end{split}
\end{equation}
where the samples of all classes take the same way to calculate categorical probability. As described in Figure~\ref{fig:analysis}(c), previous samples of old classes acquire some advantages in the propagation of gradient. Not only the proxies of old classes obtain more positive gradient, but also the proxies of new classes receive more negative gradient. Although the phenomenon of CF can be alleviated to some extent, its effect is still limited. Since the number of samples for each class in the fixed buffer will decrease as the learning process goes on, the gradient of old classes are not enough.

\textbf{SS-IL}~\cite{ahn2021ss} separately calculates categorical probability for old and new classes by separated softmax as 
\begin{equation}\small
    \label{eq:ssilloss}
\begin{split}
    L_{SS}=E_{(\bm{x},y)\sim{\mathcal{D}_t}}[-log(\frac{e^{\langle h(\bm{x};\bm{\Phi}),\bm{w}_y\rangle\cdot\gamma}}{\sum_{j\in\boxed{\mathcal{C}_{t}}}e^{\langle h(\bm{x};\bm{\Phi}),\bm{w}_j\rangle\cdot\gamma}})]\\
	+E_{(\bm{x},y)\sim{\mathcal{M}}}[-log(\frac{e^{\langle h(\bm{x};\bm{\Phi}),\bm{w}_y\rangle\cdot\gamma}}{\sum_{j\in\boxed{\mathcal{C}_{1:t-1}}}e^{\langle h(\bm{x};\bm{\Phi}),\bm{w}_j\rangle\cdot\gamma}})].
\end{split}
\end{equation}
As demonstrated in Figure~\ref{fig:analysis}(d), it cuts off the propagation from the learning of old classes to the proxies of new classes, and prevents the propagation from the learning of new classes to the proxies of old classes. It is able to avoid that the gradient of new classes affect the proxies of old classes. However, the model can not well distinguish new classes from old classes, since the lacking of gradient makes it difficult for the model to classify classes across tasks.

\textbf{ER-ACE}~\cite{caccia2021new} is also proposed to address the same issue by an asymmetric cross-entropy loss, which is expressed as 
\begin{equation}\small
    \label{eq:aceloss}
\begin{split}
    L_{ACE}=E_{(\bm{x},y)\sim{\mathcal{D}_t}}[-log(\frac{e^{\langle h(\bm{x};\bm{\Phi}),\bm{w}_y\rangle\cdot\gamma}}{\sum_{j\in\boxed{\mathcal{C}_{t}}}e^{\langle h(\bm{x};\bm{\Phi}),\bm{w}_j\rangle\cdot\gamma}})]\\
	+E_{(\bm{x},y)\sim{\mathcal{M}}}[-log(\frac{e^{\langle h(\bm{x};\bm{\Phi}),\bm{w}_y\rangle\cdot\gamma}}{\sum_{j\in\boxed{\mathcal{C}_{1:t}}}e^{\langle h(\bm{x};\bm{\Phi}),\bm{w}_j\rangle\cdot\gamma}})].
\end{split}
\end{equation}
Its categorical probability of new classes is similar with SS-IL, and the categorical probability of old classes is the same as ER. In detail, it only selects part of anchor-to-proxy pairs for the learning of new classes. As shown in Figure~\ref{fig:analysis}(e), it only breaks the gradient propagation from the learning of new classes to the proxies of old classes. Keeping the gradient from the learning of old classes to the proxies of new classes helps to avoid the inseparable situation of SS-IL. Although it is beneficial for old classes, the performance on new classes is harmed.

\subsection{Analysis of Contrastive-based Manner}
SCR~\cite{mai2021supervised} is proposed as a good alternative for online CICL by contrastive-based loss, which is denoted as
\begin{equation}\small
\begin{split}
    \label{eq:scrloss}  
    & L_{SCR}=E_{(\bm{x},y)\sim{\mathcal{D}_t}\cup\mathcal{M}}\\
    & [\frac{-1}{|P(\bm{x})|}\sum_{p\in P(\bm{x})}
    log\frac{e^{\langle h(\bm{x};\bm{\Phi}),h(\bm{x}_p;\bm{\Phi})\rangle/\mathcal{T}}}{\sum_{j\in \boxed{J(\bm{x})}}e^{\langle h(\bm{x};\bm{\Phi}),h(\bm{x}_j;\bm{\Phi})\rangle/\mathcal{T}}}].
\end{split}
\end{equation} 
It splices current samples and previous samples into the same batch and calculates the similarities of anchor-to-samples pairs. $J(\bm{x})$ is the indices set of samples except for anchor $\bm{x}$ in the same batch, while $P(\bm{x})$ denotes the set of samples with the same labels as anchor $\bm{x}$. Different from proxy-based loss, the selected pairs do not rely on the number of classes, but are related to the number of samples in a training batch. Hence, its effect is constrained by the size of memory buffer and batch size. And its performance would be not satisfactory when less samples to replay.

\section{Methodology}
\subsection{Motivation}
From above analysis, we can draw three conclusions. First and foremost, the unbalanced gradient propagation between new classes and old classes is the main cause of CF. The new classes dominate this process, making the samples of new classes highly distinguishable but the ones of old classes indivisible. Effectively controlling the gradient propagation between old and new classes can help the model alleviate the forgetting problem. Second, existing proxy-based approaches control the gradient propagation by selecting part of anchor-to-proxy pairs to calculate the objective function. Although they are effective, they are easy to hurt the generalization ability of model to learn new classes. Finally, the contrastive-based manner depends on the samples from the same batch but lacks the support of proxies. Its selection of anchor-to-sample pairs provides a heuristic way to select anchor-to-proxy pairs.

Based on these conclusions, we find that the coupling of these two manners would lead to a better solution. To avoid the limit caused by the size of samples, we do not take the coupling method in\cite{yao2022pcl}, which adds anchor-to-sample pairs to anchor-to-proxy pairs in cross-entropy loss. Specifically, we replace the samples of anchor-to-sample pairs by proxies for in contrastive-based loss, and obtain our manner

\begin{equation}\small
\begin{split}
    \label{eq:oursloss}
    & L_{ours}=E_{(\bm{x},y)\sim{\mathcal{D}_t}\cup\mathcal{M}}
    \\
    & [\frac{-1}{|P(\bm{x})|}
    \sum_{p\in P(\bm{x})}
    log\frac{e^{\langle h(\bm{x};\bm{\Phi}), \bm{w}_p\rangle/\mathcal{T}}}{\sum_{j\in \boxed{J(\bm{x})}}e^{\langle h(\bm{x};\bm{\Phi}),\bm{w}_j\rangle/\mathcal{T}}}].
\end{split}
\end{equation}

Different from existing studies, its way of computing categorical probability is changed for each mini-batch. On the one hand, such a loss has faster convergence speed and better robustness, and can cope with a small number of samples with the help of proxies. On the other hand, the replacing proxies are only from the classes that appear in the training batch. As a result, the gradient for propagation are only from the learning of these classes. As shown in Figure~\ref{fig:analysis}(f), the gradient among all proxies are not completely separated in the whole training process. The gradient propagation only occurs when the corresponding classes appear in the same batch. Meanwhile, in each learning step, only new and old classes in current batch participate in the gradient propagation. The proxies of old classes, which are affected by the negative gradient of new classes, can also generate the positive gradient for confrontation and further mitigate the forgetting problem. Hence, the samples of all classes can be recognized more correctly than existing methods.

\subsection{Proxy-based Contrastive Replay}
\label{sec:method}
With these inspirations, we propose a novel proxy-based contrastive replay (PCR) framework, and the technical details will be stated in this section. The framework consists of a CNN-based backbone $h(\bm{x};\bm{\Phi})$ and a proxy-based classifier $f(\bm{z};\bm{W})$. The whole training and inference procedures of PCR are summarized in Algorithm~\ref{alg:pcr}.

\subsubsection{The Training Procedure of PCR}
In this part, the model is trained by learning samples of new classes and replaying samples of old classes. For each task, given current samples $(\bm{x}_c, y_c)$, it randomly retrieves previous samples $(\bm{x}_\mathcal{M}, y_\mathcal{M})$ from the memory buffer (line 1-4). Besides, these original samples and their augmented samples are spliced together for the batch of training (line 5-7). Then, the model is optimized by this training batch (line 8-9). The objective function is defined as 
\begin{equation}\small
    \label{eq:pcrloss}
\begin{split}
    L_{PCR}=E_{(\bm{x},y)\sim{\mathcal{D}_t\cup\mathcal{M}}}[-log(\frac{e^{\langle h(\bm{x};\bm{\Phi}),\bm{w}_y\rangle\cdot\gamma}}{\sum_{j\in\boxed{\mathcal{C}_{B}}}e^{\langle h(\bm{x};\bm{\Phi}),\bm{w}_j\rangle\cdot\gamma}})],
\end{split}
\end{equation}
where $\mathcal{C}_{B}$ is the classes indices in current batch of training, and the indices can be repeated. Finally, it updates the memory buffer by reservoir sampling strategy, which can ensure that the probability of each sample being extracted is equal. Conveniently, the memory buffer in our framework has a fix-sized, no matter how large the amount of samples is.

\begin{algorithm}[t]\small
\caption{Proxy-based Contrastive Replay}
\label{alg:pcr}
\textbf{Input}:Dataset $D$, Learning Rate $\lambda$, Scale factor $\gamma$\\
\textbf{Output}:Network Parameters $\theta$\\
\textbf{Initialize}:Memory Buffer $\mathcal{M}\leftarrow\{\}$, Network Parameters $\bm{\theta}=\{\bm{\Phi},\bm{W}\}$
\begin{algorithmic}[1]
\FOR{$t\in\{1,2,...,T\}$}
\STATE //$Training\ Procedure$
    \FOR{mini-batch $(\bm{x}_c, y_c)\sim D_t$}
        \STATE$(\bm{x}_\mathcal{M}, y_\mathcal{M})\leftarrow RandomRetrieval(\mathcal{M})$.
        \STATE$(\bm{x}_{ori}, y_{ori})\leftarrow Concat([(\bm{x}_c, y_c),(\bm{x}_\mathcal{M}, y_\mathcal{M})])$.
        \STATE$(\bm{x}_{aug}, y_{aug})\leftarrow DataAugmentation(\bm{x}_{ori}, y_{ori})$.
        \STATE$(\bm{x}, y)\leftarrow Concat([(\bm{x}_{ori}, y_{ori}),(\bm{x}_{aug}, y_{aug})])$.
        \STATE$L=-log(\frac{e^{\langle h(\bm{x};\bm{\Phi}),\bm{w}_y\rangle\cdot\gamma}}{\sum_{j\in\mathcal{C}_{B}}e^{\langle h(\bm{x};\bm{\Phi}),\bm{w}_j\rangle\cdot\gamma}})$
        \STATE$\theta\leftarrow\theta+\lambda\nabla_{\theta}L$.
        \STATE$\mathcal{M}\leftarrow ReservoirUpdate(\mathcal{M},(\bm{x}_t, y_t))$.
    \ENDFOR
\STATE //$Inference\ Procedure$
    \FOR{$k\in\{1,2,...,m\}$}
    \STATE $y_k^*\leftarrow= \mathop{\arg\max}_{c}\frac{e^{\langle h(\bm{x}_k;\bm{\Phi}),\bm{w}_c\rangle\cdot\gamma}}{\sum_{j\in\mathcal{C}_{1:t}}e^{\langle h(\bm{x}_k;\bm{\Phi}),\bm{w}_j\rangle\cdot\gamma}},c\in C_{1:t}$
    \ENDFOR
\STATE \textbf{return} $\theta$
\ENDFOR
\end{algorithmic}
\end{algorithm}

\subsubsection{The Inference Procedure of PCR}
The inference procedure (line 13-15) is different from the training procedure. Each testing sample $\bm{x}_k$ obtains its class probability distribution by Equation (\ref{eq:probability}). And we perform the inference prediction to $\bm{x}_k$ with highest probability as
\begin{equation}\small
\small
 	\begin{split}
	 	y_k^* = \mathop{\arg\max}_{c}\frac{e^{\langle h(\bm{x}_k;\bm{\Phi}),\bm{w}_c\rangle\cdot\gamma}}{\sum_{j\in\boxed{\mathcal{C}_{1:t}}}e^{\langle h(\bm{x}_k;\bm{\Phi}),\bm{w}_j\rangle\cdot\gamma}},c\in C_{1:t}.
 	\end{split}
\end{equation}

\section{Performance Evaluation}
\label{sec:experiments}

\subsection{Experiment Setup}
\subsubsection{Datasets}
We conduct experiments on three real-world image datasets for evaluation. Split CIFAR10~\cite{krizhevsky2009learning} is split into 5 tasks, and each task contains 2 classes. Split CIFAR100~\cite{krizhevsky2009learning} as well as Split MiniImageNet~\cite{vinyals2016matching} are organized into 10 tasks, and each task is made up of samples from 10 classes.

\subsubsection{Evaluated Baselines}
To evaluate the effectiveness of PCR, we compare it with the following four methodological categories. \textbf{None-replay operations} contain IID and FINE-TUNE. \textbf{Memory update strategies} include ER~\cite{rolnick2019experience}, GSS~\cite{aljundi2019gradient}, and GMED~\cite{jin2021gradient}. MIR~\cite{aljundi2019online} and ASER~\cite{shim2021online} are \textbf{memory retrieval strategies}. And \textbf{model update strategies} contains A-GEM~\cite{chaudhry2018efficient}, ER-WA~\cite{zhao2020maintaining}, DER++~\cite{buzzega2020dark}, SS-IL~\cite{ahn2021ss}, SCR~\cite{mai2021supervised}, ER-ACE~\cite{caccia2021new}, ER-DVC~\cite{gu2022not}, and OCM~\cite{guo2022online}.

\begin{table*}[t]\small
\renewcommand\tabcolsep{2pt}
\centering
\caption{Final Accuracy Rate (higher is better). The best scores are in boldface, and the second best scores are underlined.}
\vspace{-1ex}
\label{tableaccuracy}
\begin{tabular}{l|cccc|cccc|cccc}
\hline
Datasets [sample size] & \multicolumn{4}{c|}{Split CIFAR10 [32$\times$32]} & \multicolumn{4}{c|}{Split CIFAR100 [32$\times$32]}& \multicolumn{4}{c}{Split MiniImageNet [84$\times$84]}\\ \hline
Buffer & \multicolumn{1}{c|}{100} & \multicolumn{1}{c|}{200} & \multicolumn{1}{c|}{500} & \multicolumn{1}{c|}{1000}    & \multicolumn{1}{c|}{500} & \multicolumn{1}{c|}{1000} & \multicolumn{1}{c|}{2000} & \multicolumn{1}{c|}{5000}     &\multicolumn{1}{c|}{500} & \multicolumn{1}{c|}{1000} & \multicolumn{1}{c|}{2000} & 5000     \\ \hline
IID          & \multicolumn{4}{c|}{55.9\scriptsize±0.4}                                  & \multicolumn{4}{c|}{17.1\scriptsize±1.0}                                     & \multicolumn{4}{c}{17.3\scriptsize±1.7}                                     \\
FINE-TUNE          & \multicolumn{4}{c|}{17.9\scriptsize±0.4}                                  & \multicolumn{4}{c|}{5.9\scriptsize±0.2}                                     & \multicolumn{4}{c}{4.3\scriptsize±0.2}                                     \\\hline

ER & 33.8\scriptsize±3.2& 41.7\scriptsize±2.8                 & 46.0\scriptsize±3.5                 & 46.1\scriptsize±4.3 &14.5\scriptsize±0.8& 17.6\scriptsize±0.9                  & 19.7\scriptsize±1.6                  & 20.9\scriptsize±1.2 & 11.2\scriptsize±0.6 & 13.4\scriptsize±0.9                   & 16.5\scriptsize±0.9                  & 16.2\scriptsize±1.7\\
GSS & 23.1\scriptsize±3.9& 28.3\scriptsize±4.6                 & 36.3\scriptsize±4.1                 & 44.8\scriptsize±3.6 & 14.6\scriptsize±1.3& 16.9\scriptsize±1.4                  & 19.0\scriptsize±1.8                  & 20.1\scriptsize±1.1 & 10.3\scriptsize±1.5 & 13.9\scriptsize±1.0                   & 14.6\scriptsize±1.1                  & 15.5\scriptsize±0.9\\
GMED (NeurIPS2021) & 32.8\scriptsize±4.7& 43.6\scriptsize±5.1                 & 52.5\scriptsize±3.9                 & 51.3\scriptsize±3.6 & 15.0\scriptsize±0.9& 18.8\scriptsize±0.7                  & 21.1\scriptsize±1.2                  & 23.0\scriptsize±1.5 & 11.9\scriptsize±1.2 & 15.3\scriptsize±1.3                   & 18.0\scriptsize±0.8                  & 19.6\scriptsize±1.0\\
MIR (NeurIPS2019)& 34.8\scriptsize±3.3& 40.3\scriptsize±3.3                 & 42.6\scriptsize±1.7                 & 47.4\scriptsize±4.1 & 14.8\scriptsize±0.7& 18.1\scriptsize±0.7                  & 20.3\scriptsize±1.6                  & 21.6\scriptsize±1.7 & 11.9\scriptsize±0.6 & 14.8\scriptsize±1.1                   & 17.2\scriptsize±0.8                  & 17.2\scriptsize±1.2\\
ASER (AAAI2021) & 33.7\scriptsize±3.7& 31.6\scriptsize±3.4                 & 42.1\scriptsize±3.0                 & 42.3\scriptsize±2.9 & 13.0\scriptsize±0.9& 16.1\scriptsize±1.1                  & 17.7\scriptsize±0.7                  & 18.9\scriptsize±1.0 & 10.5\scriptsize±1.1 & 13.8\scriptsize±0.9                   & 16.1\scriptsize±0.9                  & 18.1\scriptsize±1.1\\
A-GEM (ICLR2019)& 17.5\scriptsize±1.7& 17.4\scriptsize±2.1                 & 17.9\scriptsize±0.7                 & 18.2\scriptsize±1.5 & 5.4\scriptsize±0.6& 5.6\scriptsize±0.5                  & 5.4\scriptsize±0.7                  & 4.6\scriptsize±1.0 & 5.0\scriptsize±1.0 & 4.7\scriptsize±1.1                   & 5.0\scriptsize±2.3                  & 4.8\scriptsize±0.8\\
ER-WA (CVPR2020)& 36.9\scriptsize±2.9& 42.5\scriptsize±3.4                 & 48.6\scriptsize±2.7                 & 45.9\scriptsize±5.3 & 18.3\scriptsize±0.7& 21.7\scriptsize±1.2                  & 23.6\scriptsize±0.9                  & 24.0\scriptsize±1.8 & 15.1\scriptsize±0.7 & 17.1\scriptsize±0.9                   & 18.9\scriptsize±1.4                  & 18.5\scriptsize±1.5\\
DER++ (NeurIPS2020)& 40.9\scriptsize±1.4& 45.3\scriptsize±1.7                 & 52.8\scriptsize±2.2                 & 53.9\scriptsize±1.9 & 15.5\scriptsize±1.0& 17.2\scriptsize±1.1                  & 19.5\scriptsize±1.2                  & 20.2\scriptsize±1.3 & 11.9\scriptsize±1.0 & 14.8\scriptsize±0.7                   & 16.1\scriptsize±1.3                  & 15.5\scriptsize±1.3\\
SS-IL (ICCV2021) & 36.8\scriptsize±2.1 & 42.2\scriptsize±1.4                 &  44.8\scriptsize±1.6                 & 47.4\scriptsize±1.5 & 19.5\scriptsize±0.6  & 21.9\scriptsize±1.1                  & 24.5\scriptsize±1.4                  & 24.7\scriptsize±1.0 & 18.0\scriptsize±0.7 & 19.7\scriptsize±0.9                   & 21.7\scriptsize±1.0                  & 24.4\scriptsize±1.6 \\
SCR (CVPR-W2021) & 35.0\scriptsize±2.9& 45.4\scriptsize±1.0                 & 55.7\scriptsize±1.6                 & \textbf{59.8\scriptsize±1.6} & 13.3\scriptsize±0.6& 16.2\scriptsize±1.3                  & 18.2\scriptsize±0.8                  & 19.3\scriptsize±1.0 & 12.1\scriptsize±0.7 & 14.7\scriptsize±1.9                   & 16.8\scriptsize±0.6                  & 18.6\scriptsize±0.5\\
ER-DVC (CVPR2022) & 36.3\scriptsize±2.6 & 45.4\scriptsize±1.4                 &  50.6\scriptsize±2.9                & 52.1\scriptsize±2.5 & 16.8\scriptsize±0.8 & 19.7\scriptsize±0.7                  & 22.1\scriptsize±0.9                  & 24.1\scriptsize±0.8 & 13.9\scriptsize±0.6 & 15.4\scriptsize±0.7                   & 17.2\scriptsize±0.8                  & 19.1\scriptsize±0.9 \\
OCM (ICML2022) & \underline{44.4\scriptsize±1.5} & \underline{49.9\scriptsize±1.8}                 &  \underline{55.8\scriptsize±2.3}                & \underline{59.2\scriptsize±2.2} & 17.7\scriptsize±1.0 & 20.6\scriptsize±1.2                  & 22.1\scriptsize±1.0                  & 22.7\scriptsize±1.4 & 11.1\scriptsize±0.6 & 13.6\scriptsize±0.7                   & 16.5\scriptsize±0.5                  & 19.2\scriptsize±0.7 \\
ER-ACE (ICLR2022)   & 44.3\scriptsize±1.5 & 49.7\scriptsize±2.4                & 54.9\scriptsize±1.4                 & 57.5\scriptsize±1.9& \underline{19.7\scriptsize±0.8} & \underline{23.1\scriptsize±0.8}                &\underline{24.8\scriptsize±0.9}                  & \underline{27.0\scriptsize±1.2} &  \underline{18.1\scriptsize±0.5}& \underline{20.3\scriptsize±1.3}                   & \underline{24.8\scriptsize±1.1}                  & \underline{26.2\scriptsize±1.0} \\
$\hookrightarrow$ER-ACE-NCM   & \scriptsize(45.0±1.3)  & \scriptsize(51.0±1.2)  & \scriptsize(56.8±1.1)  & \scriptsize(60.1±1.0) & \scriptsize(21.0±0.6)  & \scriptsize(24.2±0.7)  & \scriptsize(26.6±0.7)  & \scriptsize(29.1±1.0)   & \scriptsize(18.1±0.8)  & \scriptsize(22.3±0.5)  & \scriptsize(25.3±0.5)  & \scriptsize(27.1±0.9) \\
\hline
PCR (Ours)   & \textbf{45.4\scriptsize±1.3}& \textbf{50.3\scriptsize±1.5}                & \textbf{56.0\scriptsize±1.2}                 & 58.8\scriptsize±1.6 & \textbf{21.8\scriptsize±0.9} & \textbf{25.6\scriptsize±0.6}                &\textbf{27.4\scriptsize±0.6}                 & \textbf{29.3\scriptsize±1.1} &  \textbf{20.9\scriptsize±0.9} & \textbf{24.2\scriptsize±0.9}                   & \textbf{27.2\scriptsize±1.2}                  & \textbf{28.4\scriptsize±0.9} \\
$\hookrightarrow$PCR-NCM   & \scriptsize(43.7±1.2)& \scriptsize(49.1±1.3)                & \scriptsize(56.2±1.2)                 & \scriptsize(59.9±1.8) & \scriptsize(22.6±0.6) & \scriptsize(26.0±0.4)                &\scriptsize(28.2±0.6)                  & \scriptsize(30.1±1.0) &  \scriptsize(19.8±0.6)& \scriptsize(23.5±0.6)                  & \scriptsize(26.8±0.5)                  & \scriptsize(28.0±0.6) \\
\hline
\end{tabular}
\vspace{-3ex}
\end{table*}

\subsubsection{Evaluation Metrics}
We need to measure the performance of model for online CICL. Most important of all, we define $a_{i,j}(j<=i)$ as the accuracy evaluated on the held-out test samples of the $j$th task after the network has learned the training samples in the first $i$ tasks. Similar with~\cite{shim2021online}, we can acquire average accuracy rate$A_i$ at the $i$th task based on $a_{i,j}(j<=i)$. 
\begin{equation}
\label{eq:first}
 	\begin{split}
	 	A_i = \frac{1}{i}\sum_{j=1}^{i}a_{i,j}
 	\end{split}
\end{equation}
If the model finish learning all of $T$ tasks, $A_T$ is equivalent to the final accuracy rate. Furthermore, we decompose the accuracy rate to obtain the average accuracy rate of new data $A^n_i=a_{i,i}$ and the one of old data $A^o_i$ at the $i$th task, where

\begin{equation}
\label{eq:third}
 	\begin{split}
	 	A^o_i = \frac{1}{i-1}\sum_{j=1}^{i-1}a_{i,j}.
 	\end{split}
\end{equation}

\subsubsection{Implementation Details}

The basic setting of backbone model is the same as the latest work~\cite{caccia2021new}. In detail, we take the Reduced ResNet18 (the number of filters is 20) as the feature extractor for all datasets. During the training phase, the network is trained with the SGD optimizer and the learning rate is set as 0.1. 

For all datasets, the classes are shuffled before division. And we set the memory buffer with \{100, 200, 500, 1000\} for Split CIFAR10, and with \{500, 1000, 2000, 5000\} for other two datasets. The model receives 10 current samples from data stream and 10 previous samples from the memory buffer at a time irrespective of the size of the memory. Moreover, we employ a combination of various augmentation techniques to get the augmented images. And the usage of data augmentation is fair for all methods. As for the testing phase, we set 256 as the batch size of validation.

\begin{figure*}[t]
\centering
    \begin{minipage}[t]{0.44\linewidth}
        \centering
        \includegraphics[scale=0.21]{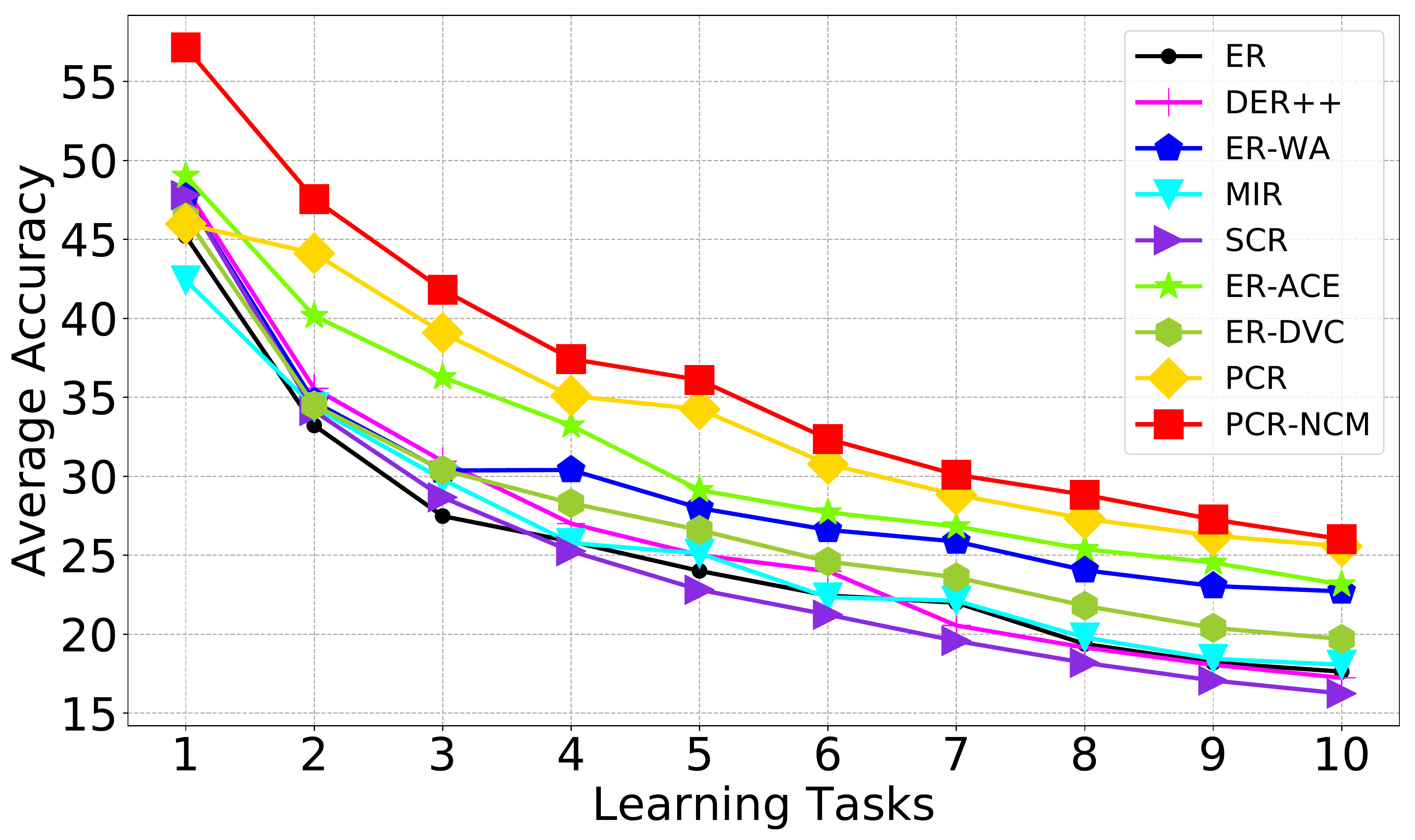}
        {(a) Split CIFAR100}
    \end{minipage}
    \begin{minipage}[t]{0.44\linewidth}
        \centering
        \includegraphics[scale=0.21]{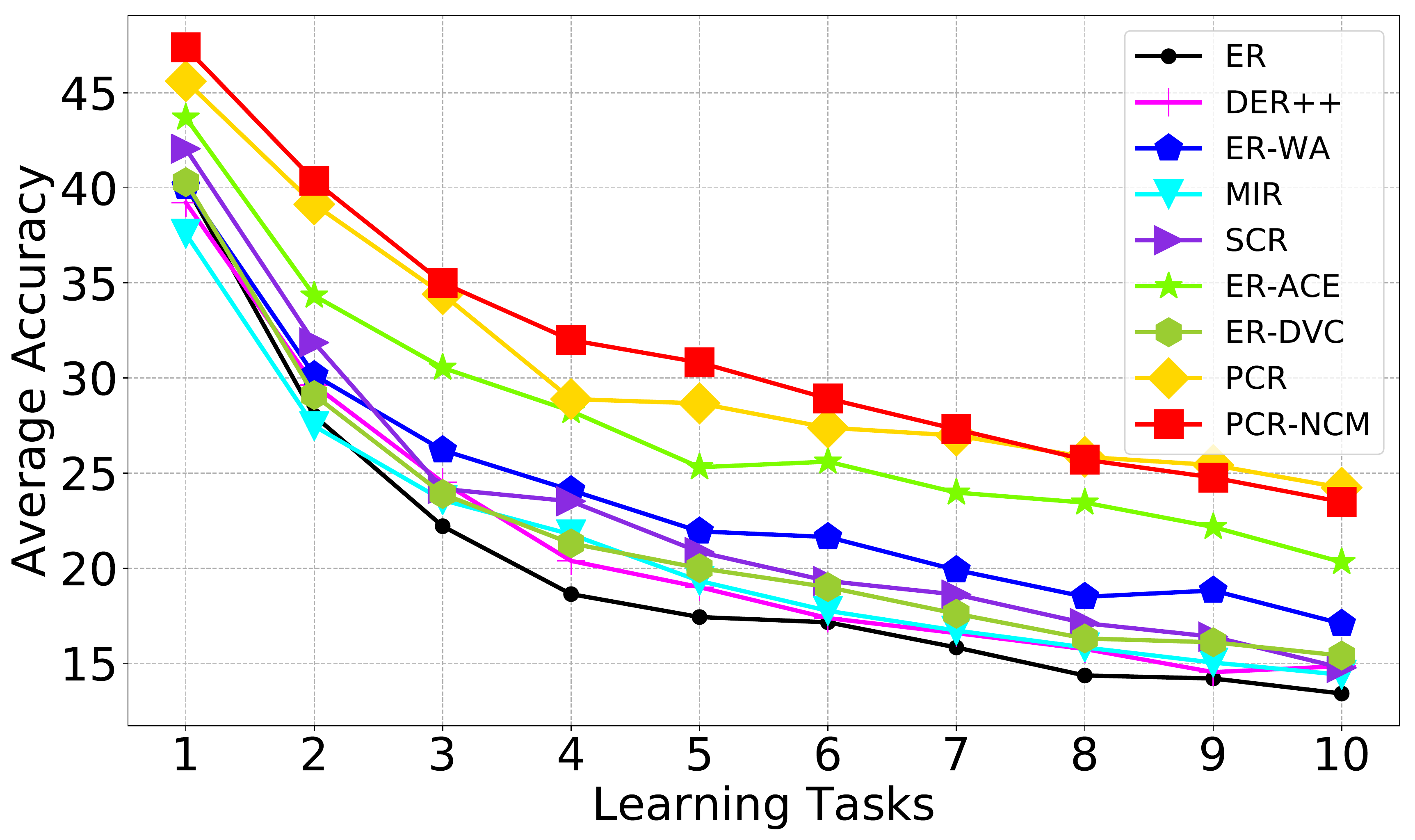}
        {(b) Split MiniImageNet}
    \end{minipage}
\centering
\vspace{-2ex}
\caption{Average accuracy rate on observed learning stages on Split CIFAR100 and Split MiniImageNet while the buffer size is 1000. }
\label{learningstage1000}
\vspace{-3ex}
\end{figure*}

\begin{table}[t]
\centering
\caption{Final Accuracy Rate (higher is better) on Split CIFAR100.}
\vspace{-1ex}
\label{moreaccuracy}
\begin{tabular}{l|lll}
\hline
Buffer  & 1000       & 2000      & 5000      \\ \hline
SCR~\cite{guo2022online}     & 26.5\scriptsize±0.2           & 31.6\scriptsize±0.5          & 36.5\scriptsize±0.2          \\
OCM~\cite{guo2022online}      & 28.1\scriptsize±0.3           & 35.0\scriptsize±0.4          & 42.4\scriptsize±0.5          \\
PCR     & 29.3\scriptsize±0.6           & 36.3\scriptsize±0.9          & 46.5\scriptsize±0.8          \\
PCR-NCM     & 31.0\scriptsize±0.9           & 37.7\scriptsize±0.8          & 47.9\scriptsize±0.6          \\\hline
\end{tabular}
\vspace{-3ex}
\end{table}

\begin{figure*}[t]
\centering
    \begin{minipage}[t]{0.44\linewidth}
        \centering
        \includegraphics[scale=0.21]{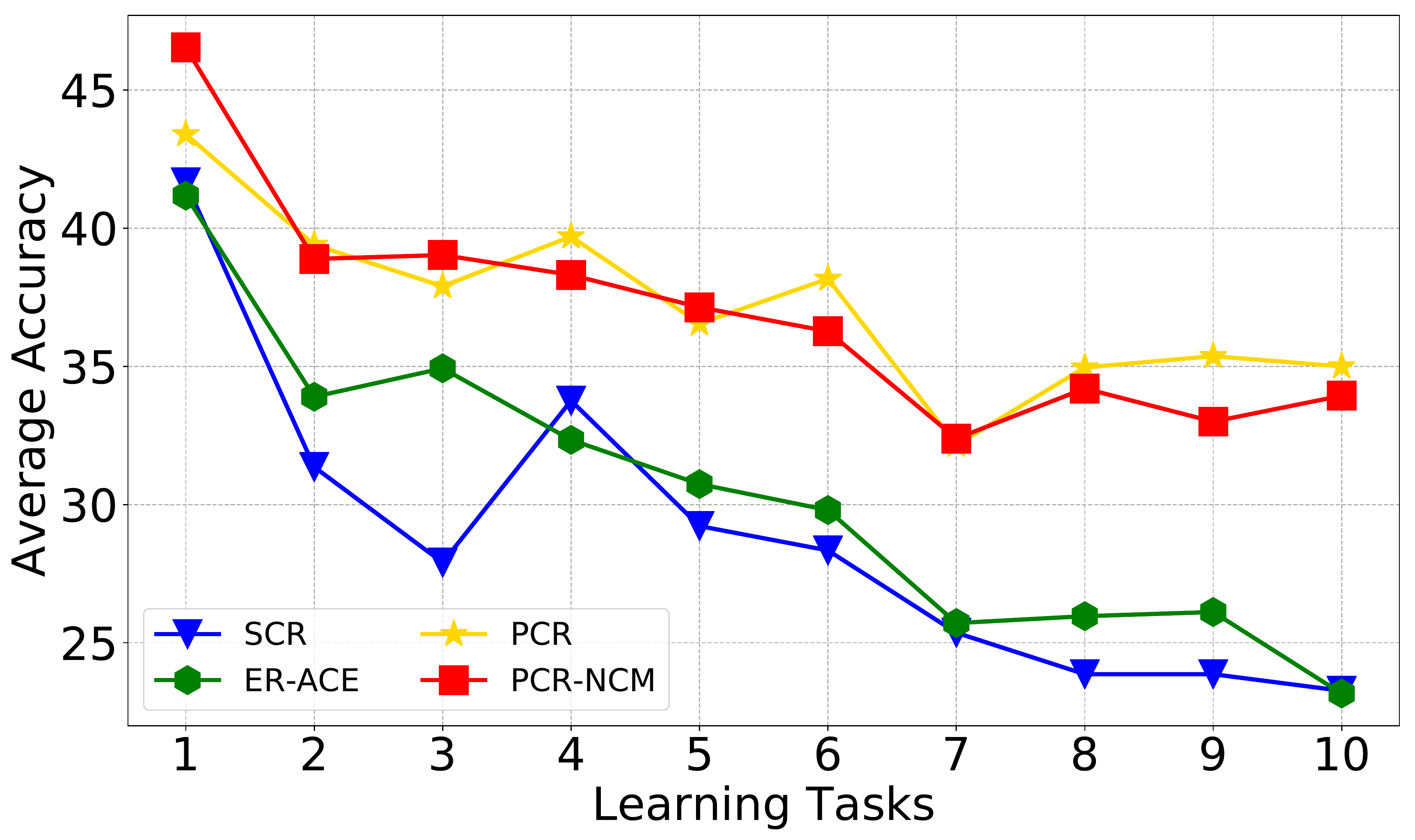}
        {(a) Performance on novel knowledge}
    \end{minipage}
    \begin{minipage}[t]{0.44\linewidth}
        \centering
        \includegraphics[scale=0.21]{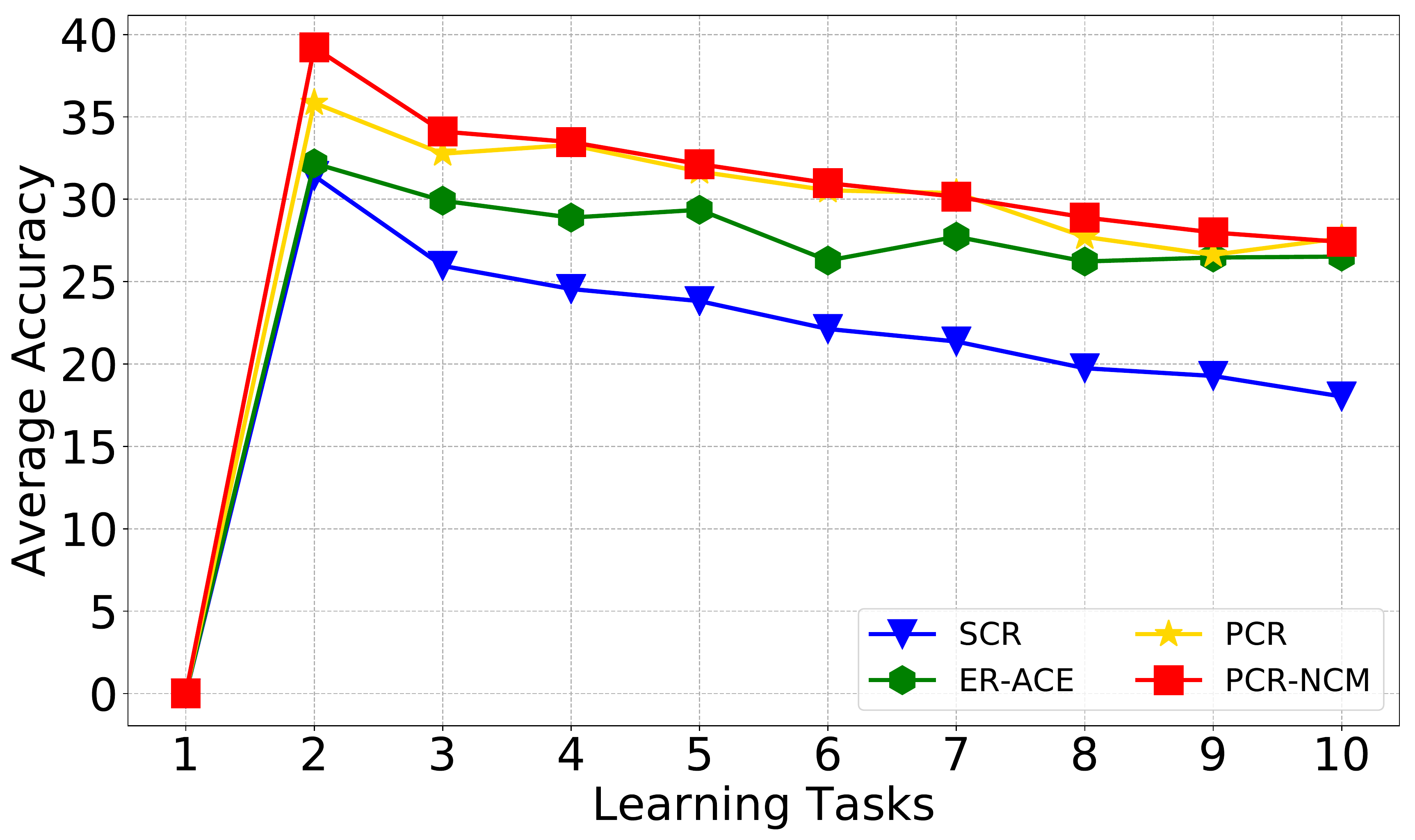}
        {(b) Performance on historical knowledge}
    \end{minipage}
\centering
\vspace{-2ex}
\caption{Average accuracy rate on observed learning stages Split MiniImageNet while the buffer size is 5000. }
\label{learningstage5000}
\vspace{-3ex}
\end{figure*}

\subsection{Overall Performance}
In this section, we conduct experiments to compare with various state-of-the-art baselines of continual learning. Table~\ref{tableaccuracy} shows the final average accuracy rate for Split CIFAR10, Split CIFAR100 and Split MiniImageNet. All reported scores are the average score of 10 runs with the 95\%
confidence interval. It is easy to find that our method obtains significantly improved performance on three datasets.

\textbf{Comparison on final accuracy.} Table~\ref{tableaccuracy} reports the accuracy performance of all baseline methods on three datasets. By comparing the final accuracy of all methods, we can draw two conclusions. First, the model update strategies are more effective and can greatly improve the performance among all replay-based methods. Second, ER-ACE achieves the highest accuracy rate as the latest method. 

Our method PCR achieves the best performance, which confirms its effectiveness. In general, PCR achieves the best performance under most experimental settings in which each dataset contains four memory buffer of different sizes. On the one hand, it has the most outstanding performance on Split MiniImageNet among three benchmarks. On the other hand, the growth of buffer size further improve the performance of PCR. For example, PCR improves the proxy-based baseline ER and contrastive-based baseline SCR with a gap of more than 10\%. Meanwhile, PCR outperforms the strongest baseline ER-ACE by 2.8\%, 3.9\%, 2.4\%, 2.2\% on Split MiniImagenet when the size of memory buffer is 500, 1000, 2000 and 5000, respectively. In addition, PCR defeats ER-ACE with an improvement of 2.1\%, 2.5\%, 2.6\% and 2.3\% on Split CIFAR100 with 500, 1000, 2000 and 5000 size of memory buffer, respectively. 

Although there is a scenario on Split CIFAR10 that does not achieve the strongest result, the overall performance on this dataset is the best. By the buffer with 1000 samples, SCR that uses NCM classifier beats PCR on Split CIFAR10. We replace the proxy-based classifier by the NCM one and get the ER-ACE-NCM as well as PCR-NCM. By NCM classifier, there will be some significant improvement in their performance. And the results state that NCM classifier is more suitable for situation with smaller samples (e.g. Split-CIFAR100) and a larger buffer. Since small-sized samples are easy to obtain reliable features, and large-sized buffer trends to obtain accurate classification centers. In addition, OCM performs better than PCR when the buffer is large with the advantages of multiple data augmentations. 

In the same time, we also compare PCR with SCR and OCM under the experiment setting in \cite{guo2022online}, shown as Table~\ref{moreaccuracy}. In~\cite{guo2022online}, the model is set as ResNet18 (the number of filters is 64), and it retrieves 64 samples from memory buffer for each training batch. Moreover, it is trained with Adam optimizer and the learning rate is set as 0.001. These experimental conditions are different from ours and can significantly improve the performance. The results suggest that our method is significantly better than SCR and OCM.

\begin{table*}[t]\small
\renewcommand\tabcolsep{2.5pt}
\centering
\caption{Final Accuracy Rate (higher is better) on three datasets for ablation studies. ER is a baseline method. }
\vspace{-1ex}
\label{ablation}
\begin{tabular}{l|cccc|cccc|cccc}
\hline
Datasets & \multicolumn{4}{c|}{Split CIFAR10} & \multicolumn{4}{c|}{Split CIFAR100}& \multicolumn{4}{c}{Split MiniImageNet}\\ \hline
Buffer & \multicolumn{1}{c|}{100} & \multicolumn{1}{c|}{200} & \multicolumn{1}{c|}{500} & \multicolumn{1}{c|}{1000}    & \multicolumn{1}{c|}{500} & \multicolumn{1}{c|}{1000} & \multicolumn{1}{c|}{2000} & \multicolumn{1}{c|}{5000}     &\multicolumn{1}{c|}{500} & \multicolumn{1}{c|}{1000} & \multicolumn{1}{c|}{2000} & 5000     \\ \hline
ER & 33.8\scriptsize±3.2& 41.7\scriptsize±2.8                 & 46.0\scriptsize±3.5                 & 46.1\scriptsize±4.3 &14.5\scriptsize±0.8& 17.6\scriptsize±0.9                  & 19.7\scriptsize±1.6                  & 20.9\scriptsize±1.2 & 11.2\scriptsize±0.6 & 13.4\scriptsize±0.9                   & 16.5\scriptsize±0.9                  & 16.2\scriptsize±1.7\\
ER+A   & 39.4\scriptsize±3.2& 48.8\scriptsize±1.1                & 52.3\scriptsize±1.7                 & 53.9\scriptsize±3.4 & 20.5\scriptsize±0.9 & 25.5\scriptsize±0.6                &26.0\scriptsize±0.7                  & 28.2\scriptsize±0.7 &  19.4\scriptsize±1.3& 22.8\scriptsize±0.8                   & 25.6\scriptsize±1.1                  & 27.5\scriptsize±1.0 \\

ER+A+B (PCR)    & 45.4\scriptsize±1.3& 50.3\scriptsize±1.5                & 56.0\scriptsize±1.2                 & 58.8\scriptsize±1.6 & 21.8\scriptsize±0.9 & 25.6\scriptsize±0.6                &27.4\scriptsize±0.6                 & 29.3\scriptsize±1.1 &  20.9\scriptsize±0.9 & 24.2\scriptsize±0.9                   & 27.2\scriptsize±1.2                  & 28.4\scriptsize±0.9 \\
ER+A+B+C    & 41.9\scriptsize±2.0& 48.3\scriptsize±2.4                & 55.8\scriptsize±1.2                 & 57.0\scriptsize±2.6 & 19.8\scriptsize±1.0   & 24.1\scriptsize±0.5           & 25.9\scriptsize±0.5          & 27.3\scriptsize±0.7 &  19.3\scriptsize±1.2 & 21.8\scriptsize±0.8                   & 24.5\scriptsize±0.9                  & 26.3\scriptsize±1.0 \\

\hline
\end{tabular}
\vspace{-3ex}
\end{table*}

\textbf{Comparison on learning process.} For more detailed comparison, we reveal the accuracy performance in each task for part of effective approaches on all datasets, as shown in Figure~\ref{learningstage1000}. The results of line chart show that PCR not only achieves significant results in the accuracy of final task, but also performs better than other baselines in the whole learning process. In fact, with the help of NCM classifier, the overall performance of PCR-NCM is the best one. However, it has to calculate classification centers before its inference process, which greatly reduces its efficiency. Furthermore, the performance of PCR in the first few tasks does not outperform other baselines. However, the improvement of it become more and more visible as the number of tasks increases, proving its power to overcome CF. For instance, PCR has no obvious advantages in the first task, but it shows real effect in the remaining tasks on Split CIFAR100. Meanwhile, PCR and ER-ACE, especially PCR, are far superior in performance to other baselines on Split MiniImagenet. Therefore, our approach has a stronger ability to resist forgetting in the case of less data.

\textbf{Comparison on knowledge balance.} Actually, we should not only focus on the model's ability to retain historical knowledge, but also ensure the model's ability to quickly learn novel knowledge. Shown in Figure~\ref{learningstage5000}, we record the accuracy performance of novel and historical knowledge in each task for part of effective methods on Split MiniImageNet. Although SCR and ER-ACE can improve the anti-forgetting ability of the model, they have a serious impact on the generalization ability of the model. As the historical knowledge is consolidated, the learning performance of the model on novel knowledge becomes very poor. Different from existing studies, our model can not only effectively alleviate the phenomenon of CF, but also reduce the decline of the model at the generalization level as much as possible.

\begin{figure}[t]
\centering
\includegraphics[scale=0.18]{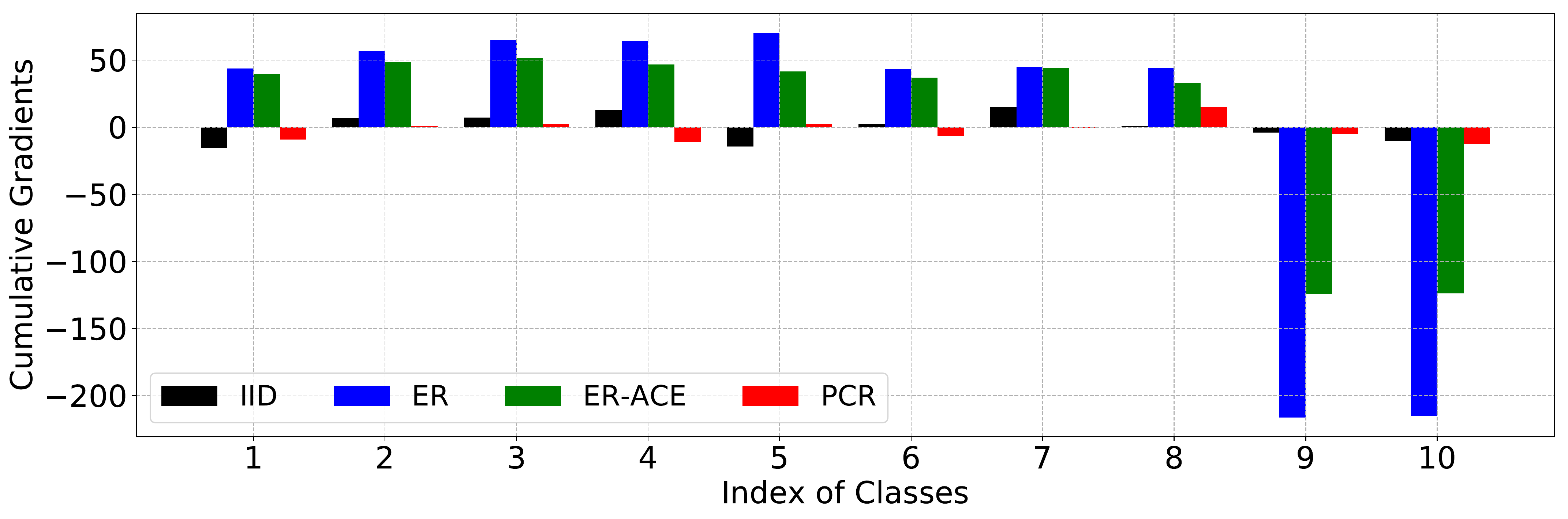}
\vspace{-2ex}
\caption{Cumulative gradients of different methods for all proxies when the model learns new classes (9/10) on Split CIFAR10.}
\label{fig:gradient}%
\vspace{-3ex}
\end{figure}

\subsection{Ablation Study}
In this section, we decompose PCR into several components, and further demonstrate their functions.

``A'' means the \textbf{selection component}, which selects the anchor-to-proxy pairs as the contrastive-based replay manner. Actually, there are some samples from the same classes in a training batch. As a result, there are some duplicate anchor-to-proxy pairs in PCR. To effectively verify the role of the selected proxies, we remove the duplicate pairs. As shown in Table~\ref{ablation}, the performance of ER has significant improvement with the help of this component.

``B'' is the \textbf{duplication component} to keep the duplicate pairs in PCR. Although it contains the same knowledge as the anchor-to-proxy pairs of selection component, it still produces significant performance. Since it can provide anchor samples with more negative pairs, which are vital to contrastive-based loss. As stated in Table~\ref{ablation}, PCR outperforms ``ER+A'' with the duplication component.

``C'' denotes the \textbf{addition component} to exploit original anchor-to-sample pairs of SCR. Combining PCR with this part, we can get another coupling manner as \cite{yao2022pcl}. It keeps anchor-to-sample pairs of contrastive-based loss, when replacing anchor-to-sample pairs by anchor-to-proxy pairs as PCR. Although it provides more knowledge about the relationships of samples, its performance is limited by little number of samples as indicated in Table~\ref{ablation}.

\begin{figure}[t]
\centering
\includegraphics[scale=0.18]{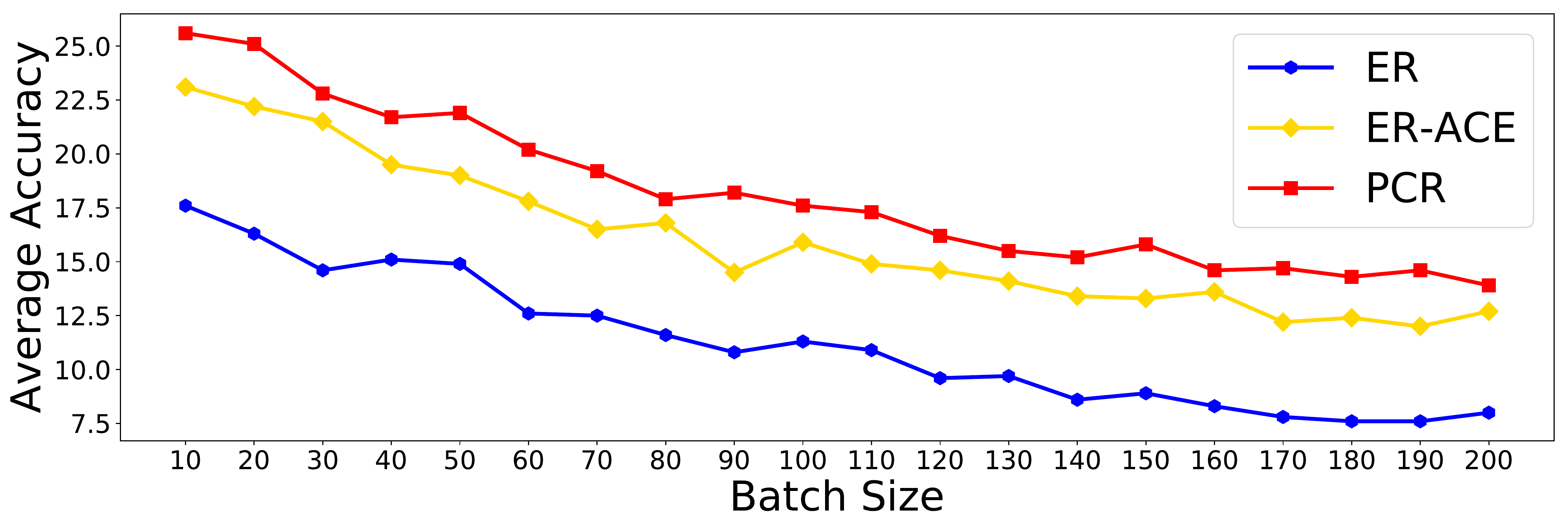}
\vspace{-2ex}
\caption{Final accuracy rate (higher is better) on Split CIFAR100 with different batch size when the buffer size is 1000.}
\label{fig:batch}%
\vspace{-3ex}
\end{figure}

In conclusion, the selection component and the duplication component are the keys of PCR. As displayed in Figure~\ref{fig:gradient}, PCR produces uniform gradients for all proxies to address the ``bias'' issue by the smart selection of anchor-to-proxy pairs. Furthermore, since the way of selection depends on the classes in the training batch, PCR is influenced by the batch size. Figure~\ref{fig:batch} reports the performance of several methods with different batch size. With the increase of batch size, PCR consistently maintains its advantages.

\section{Conclusion}
\label{sec:conclusion}
In this paper, we develop a novel online CICL method called PCR to alleviate the phenomenon of catastrophic forgetting by the coupling of proxy-based and contrastive-based replay manners. Based on the characteristics of these two manners, we propose to replace the samples of anchor-to-sample pairs by proxies. The coupling replay manner realizes complementary advantages. With the help of proxies, our method keeps the fast and stable convergence. In the meantime, the same selection of anchor-to-proxy pairs as contrastive samples is beneficial for addressing the ``bias'' issue of proxy-based manner. Extensive experiments on three datasets demonstrate the superiority of PCR over a large variety of state-of-the-art methods.

\section*{Acknowledgement}

This work was supported in part by National Nature Science Foundation of China (No.62202124 and No.62272130), Shenzhen Science and Technology Program (No.KCXFZ20211020163403005) and  Nature Science Program of Shenzhen (No.JCYJ20210324120208022 and No.JCYJ20200109113014456).

{\small


\begin{thebibliography}{10}\itemsep=-1pt

\bibitem{ahn2021ss}
Hongjoon Ahn, Jihwan Kwak, Subin Lim, Hyeonsu Bang, Hyojun Kim, and Taesup
  Moon.
\newblock Ss-il: Separated softmax for incremental learning.
\newblock In {\em Proceedings of the IEEE/CVF International Conference on
  Computer Vision}, pages 844--853, 2021.

\bibitem{aljundi2019online}
Rahaf Aljundi, Eugene Belilovsky, Tinne Tuytelaars, Laurent Charlin, Massimo
  Caccia, Min Lin, and Lucas Page-Caccia.
\newblock Online continual learning with maximal interfered retrieval.
\newblock {\em Advances in Neural Information Processing Systems},
  32:11849--11860, 2019.

\bibitem{aljundi2019gradient}
Rahaf Aljundi, Min Lin, Baptiste Goujaud, and Yoshua Bengio.
\newblock Gradient based sample selection for online continual learning.
\newblock {\em Advances in neural information processing systems}, 32, 2019.

\bibitem{buzzega2020dark}
Pietro Buzzega, Matteo Boschini, Angelo Porrello, Davide Abati, and Simone
  Calderara.
\newblock Dark experience for general continual learning: a strong, simple
  baseline.
\newblock {\em Advances in neural information processing systems},
  33:15920--15930, 2020.

\bibitem{buzzega2021rethinking}
Pietro Buzzega, Matteo Boschini, Angelo Porrello, and Simone Calderara.
\newblock Rethinking experience replay: a bag of tricks for continual learning.
\newblock In {\em 2020 25th International Conference on Pattern Recognition
  (ICPR)}, pages 2180--2187. IEEE, 2021.

\bibitem{caccia2021new}
Lucas Caccia, Rahaf Aljundi, Nader Asadi, Tinne Tuytelaars, Joelle Pineau, and
  Eugene Belilovsky.
\newblock New insights on reducing abrupt representation change in online
  continual learning.
\newblock In {\em International Conference on Learning Representations}, 2021.

\bibitem{cha2021co2l}
Hyuntak Cha, Jaeho Lee, and Jinwoo Shin.
\newblock Co2l: Contrastive continual learning.
\newblock In {\em Proceedings of the IEEE/CVF International Conference on
  Computer Vision}, pages 9516--9525, 2021.

\bibitem{chaudhry2021using}
Arslan Chaudhry, Albert Gordo, Puneet Dokania, Philip Torr, and David
  Lopez-Paz.
\newblock Using hindsight to anchor past knowledge in continual learning.
\newblock In {\em Proceedings of the AAAI Conference on Artificial
  Intelligence}, volume~35, pages 6993--7001, 2021.

\bibitem{chaudhry2018efficient}
Arslan Chaudhry, Marc’Aurelio Ranzato, Marcus Rohrbach, and Mohamed
  Elhoseiny.
\newblock Efficient lifelong learning with a-gem.
\newblock In {\em International Conference on Learning Representations}, 2018.

\bibitem{choi2021dual}
Yoojin Choi, Mostafa El-Khamy, and Jungwon Lee.
\newblock Dual-teacher class-incremental learning with data-free generative
  replay.
\newblock In {\em Proceedings of the IEEE/CVF Conference on Computer Vision and
  Pattern Recognition}, pages 3543--3552, 2021.

\bibitem{cui2021deepcollaboration}
Bo Cui, Guyue Hu, and Shan Yu.
\newblock Deepcollaboration: Collaborative generative and discriminative models
  for class incremental learning.
\newblock In {\em Proceedings of the AAAI Conference on Artificial
  Intelligence}, volume~35, pages 1175--1183, 2021.

\bibitem{delange2021continual}
Matthias Delange, Rahaf Aljundi, Marc Masana, Sarah Parisot, Xu Jia, Ales
  Leonardis, Greg Slabaugh, and Tinne Tuytelaars.
\newblock A continual learning survey: Defying forgetting in classification
  tasks.
\newblock {\em IEEE Transactions on Pattern Analysis and Machine Intelligence},
  2021.

\bibitem{dhar2019learning}
Prithviraj Dhar, Rajat~Vikram Singh, Kuan-Chuan Peng, Ziyan Wu, and Rama
  Chellappa.
\newblock Learning without memorizing.
\newblock In {\em Proceedings of the IEEE/CVF Conference on Computer Vision and
  Pattern Recognition}, pages 5138--5146, 2019.

\bibitem{farajtabar2020orthogonal}
Mehrdad Farajtabar, Navid Azizan, Alex Mott, and Ang Li.
\newblock Orthogonal gradient descent for continual learning.
\newblock In {\em International Conference on Artificial Intelligence and
  Statistics}, pages 3762--3773. PMLR, 2020.

\bibitem{gu2022not}
Yanan Gu, Xu Yang, Kun Wei, and Cheng Deng.
\newblock Not just selection, but exploration: Online class-incremental
  continual learning via dual view consistency.
\newblock In {\em Proceedings of the IEEE/CVF Conference on Computer Vision and
  Pattern Recognition}, pages 7442--7451, 2022.

\bibitem{guo2022online}
Yiduo Guo, Bing Liu, and Dongyan Zhao.
\newblock Online continual learning through mutual information maximization.
\newblock In {\em International Conference on Machine Learning}, pages
  8109--8126. PMLR, 2022.

\bibitem{he2021online}
Jiangpeng He and Fengqing Zhu.
\newblock Online continual learning for visual food classification.
\newblock In {\em Proceedings of the IEEE/CVF International Conference on
  Computer Vision}, pages 2337--2346, 2021.

\bibitem{hou2019learning}
Saihui Hou, Xinyu Pan, Chen~Change Loy, Zilei Wang, and Dahua Lin.
\newblock Learning a unified classifier incrementally via rebalancing.
\newblock In {\em Proceedings of the IEEE/CVF Conference on Computer Vision and
  Pattern Recognition}, pages 831--839, 2019.

\bibitem{jin2021gradient}
Xisen Jin, Arka Sadhu, Junyi Du, and Xiang Ren.
\newblock Gradient-based editing of memory examples for online task-free
  continual learning.
\newblock {\em Advances in Neural Information Processing Systems}, 34, 2021.

\bibitem{kirkpatrick2017overcoming}
James Kirkpatrick, Razvan Pascanu, Neil Rabinowitz, Joel Veness, Guillaume
  Desjardins, Andrei~A Rusu, Kieran Milan, John Quan, Tiago Ramalho, Agnieszka
  Grabska-Barwinska, et~al.
\newblock Overcoming catastrophic forgetting in neural networks.
\newblock {\em Proceedings of the national academy of sciences},
  114(13):3521--3526, 2017.

\bibitem{krizhevsky2009learning}
Alex Krizhevsky, Geoffrey Hinton, et~al.
\newblock Learning multiple layers of features from tiny images.
\newblock 2009.

\bibitem{lin2022anchor}
Huiwei Lin, Shanshan Feng, Xutao Li, Wentao Li, and Yunming Ye.
\newblock Anchor assisted experience replay for online class-incremental
  learning.
\newblock {\em IEEE Transactions on Circuits and Systems for Video Technology},
  2022.

\bibitem{liu2021rmm}
Yaoyao Liu, Bernt Schiele, and Qianru Sun.
\newblock Rmm: Reinforced memory management for class-incremental learning.
\newblock {\em Advances in Neural Information Processing Systems}, 34, 2021.

\bibitem{lopez2017gradient}
David Lopez-Paz and Marc'Aurelio Ranzato.
\newblock Gradient episodic memory for continual learning.
\newblock {\em Advances in neural information processing systems}, 30, 2017.

\bibitem{mai2022online}
Zheda Mai, Ruiwen Li, Jihwan Jeong, David Quispe, Hyunwoo Kim, and Scott
  Sanner.
\newblock Online continual learning in image classification: An empirical
  survey.
\newblock {\em Neurocomputing}, 469:28--51, 2022.

\bibitem{mai2021supervised}
Zheda Mai, Ruiwen Li, Hyunwoo Kim, and Scott Sanner.
\newblock Supervised contrastive replay: Revisiting the nearest class mean
  classifier in online class-incremental continual learning.
\newblock In {\em Proceedings of the IEEE/CVF Conference on Computer Vision and
  Pattern Recognition}, pages 3589--3599, 2021.

\bibitem{mensink2013distance}
Thomas Mensink, Jakob Verbeek, Florent Perronnin, and Gabriela Csurka.
\newblock Distance-based image classification: Generalizing to new classes at
  near-zero cost.
\newblock {\em IEEE transactions on pattern analysis and machine intelligence},
  35(11):2624--2637, 2013.

\bibitem{miao2021continual}
Zichen Miao, Ze Wang, Wei Chen, and Qiang Qiu.
\newblock Continual learning with filter atom swapping.
\newblock In {\em International Conference on Learning Representations}, 2021.

\bibitem{nair2010rectified}
Vinod Nair and Geoffrey~E Hinton.
\newblock Rectified linear units improve restricted boltzmann machines.
\newblock In {\em Icml}, 2010.

\bibitem{rolnick2019experience}
David Rolnick, Arun Ahuja, Jonathan Schwarz, Timothy Lillicrap, and Gregory
  Wayne.
\newblock Experience replay for continual learning.
\newblock {\em Advances in Neural Information Processing Systems}, 32, 2019.

\bibitem{rusu2016progressive}
Andrei~A Rusu, Neil~C Rabinowitz, Guillaume Desjardins, Hubert Soyer, James
  Kirkpatrick, Koray Kavukcuoglu, Razvan Pascanu, and Raia Hadsell.
\newblock Progressive neural networks.
\newblock {\em arXiv preprint arXiv:1606.04671}, 2016.

\bibitem{shim2021online}
Dongsub Shim, Zheda Mai, Jihwan Jeong, Scott Sanner, Hyunwoo Kim, and Jongseong
  Jang.
\newblock Online class-incremental continual learning with adversarial shapley
  value.
\newblock In {\em Proceedings of the AAAI Conference on Artificial
  Intelligence}, volume~35, pages 9630--9638, 2021.

\bibitem{tang2021layerwise}
Shixiang Tang, Dapeng Chen, Jinguo Zhu, Shijie Yu, and Wanli Ouyang.
\newblock Layerwise optimization by gradient decomposition for continual
  learning.
\newblock In {\em Proceedings of the IEEE/CVF Conference on Computer Vision and
  Pattern Recognition}, pages 9634--9643, 2021.

\bibitem{van2020brain}
Gido~M van~de Ven, Hava~T Siegelmann, and Andreas~S Tolias.
\newblock Brain-inspired replay for continual learning with artificial neural
  networks.
\newblock {\em Nature communications}, 11(1):1--14, 2020.

\bibitem{vinyals2016matching}
Oriol Vinyals, Charles Blundell, Timothy Lillicrap, Daan Wierstra, et~al.
\newblock Matching networks for one shot learning.
\newblock {\em Advances in neural information processing systems},
  29:3630--3638, 2016.

\bibitem{wang2021memory}
Liyuan Wang, Xingxing Zhang, Kuo Yang, Longhui Yu, Chongxuan Li, HONG Lanqing,
  Shifeng Zhang, Zhenguo Li, Yi Zhong, and Jun Zhu.
\newblock Memory replay with data compression for continual learning.
\newblock In {\em International Conference on Learning Representations}, 2021.

\bibitem{xiang2019incremental}
Ye Xiang, Ying Fu, Pan Ji, and Hua Huang.
\newblock Incremental learning using conditional adversarial networks.
\newblock In {\em Proceedings of the IEEE/CVF International Conference on
  Computer Vision}, pages 6619--6628, 2019.

\bibitem{yao2022pcl}
Xufeng Yao, Yang Bai, Xinyun Zhang, Yuechen Zhang, Qi Sun, Ran Chen, Ruiyu Li,
  and Bei Yu.
\newblock Pcl: Proxy-based contrastive learning for domain generalization.
\newblock In {\em Proceedings of the IEEE/CVF Conference on Computer Vision and
  Pattern Recognition}, pages 7097--7107, 2022.

\bibitem{yin2021mitigating}
Haiyan Yin, Ping Li, et~al.
\newblock Mitigating forgetting in online continual learning with neuron
  calibration.
\newblock {\em Advances in Neural Information Processing Systems}, 34, 2021.

\bibitem{zhao2020maintaining}
Bowen Zhao, Xi Xiao, Guojun Gan, Bin Zhang, and Shu-Tao Xia.
\newblock Maintaining discrimination and fairness in class incremental
  learning.
\newblock In {\em Proceedings of the IEEE/CVF Conference on Computer Vision and
  Pattern Recognition}, pages 13208--13217, 2020.

\bibitem{zhu2021class}
Fei Zhu, Zhen Cheng, Xu-yao Zhang, and Cheng-lin Liu.
\newblock Class-incremental learning via dual augmentation.
\newblock {\em Advances in Neural Information Processing Systems}, 34, 2021.

\end{thebibliography}
}

\end{document}